\documentclass[10pt,journal,compsoc]{IEEEtran}
\pdfoutput=1
\usepackage{soul}
\soulregister\cite7 
\soulregister\citep7 
\soulregister\citet7 
\soulregister\ref7 
\soulregister\pageref7 

\usepackage{ifpdf}
\usepackage{algorithmic}
\usepackage{algorithm}
\usepackage{amsmath}
\usepackage{amssymb}
\usepackage{array}
\usepackage{fixltx2e}
\usepackage{stfloats}
\usepackage{url}
\usepackage[colorlinks,linkcolor=red]{hyperref}

\makeatletter
\newcommand{\removelatexerror}{\let\@latex@error\@gobble}
\makeatother

\newcommand{\tabincell}[2]{\begin{tabular}{@{}#1@{}}#2\end{tabular}} 
\usepackage[table,xcdraw]{xcolor}
\definecolor{Gray}{gray}{0.9}
\newcommand\rurl[1]{%
  \href{http://#1}{\nolinkurl{#1}}%
}

\ifCLASSOPTIONcompsoc
 \usepackage[caption=false,font=footnotesize,labelfont=sf,textfont=sf]{subfig}
\else
 \usepackage[caption=false,font=footnotesize]{subfig}
\fi

\ifCLASSOPTIONcompsoc
  \usepackage[nocompress]{cite}
\else
  \usepackage{cite}
\fi

\ifCLASSINFOpdf
   \usepackage[pdftex]{graphicx}
\else
   \usepackage[dvips]{graphicx}
\fi

\begin{document}
%
\title{Adaptive Joint Optimization for 3D Reconstruction with Differentiable Rendering}
\author{
        Jingbo~Zhang, 
        Ziyu~Wan, 
        and~Jing~Liao$^*$
        
\IEEEcompsocitemizethanks{
\IEEEcompsocthanksitem $^*$: corresponding author.
\IEEEcompsocthanksitem J. Zhang, Z. Wan and J. Liao are with Department of Computer Science, City University of Hong Kong. E-mail: jbzhang6-c@my.cityu.edu.hk, ziyuwan2-c@my.cityu.edu.hk, jingliao@cityu.edu.hk.
    }
}

\IEEEtitleabstractindextext{%
\begin{abstract}
Due to inevitable noises introduced during scanning and quantization, 3D reconstruction via RGB-D sensors suffers from errors both in geometry and texture, leading to artifacts such as camera drifting, mesh distortion, texture ghosting, and blurriness. Given an imperfect reconstructed 3D model, most previous methods have focused on the refinement of either geometry, texture, or camera pose. Or different optimization schemes and objectives for optimizing each component have been used in previous joint optimization methods, forming a complicated system. In this paper, we propose a novel optimization approach based on differentiable rendering, which integrates the optimization of camera pose, geometry, and texture into a unified framework by enforcing consistency between the rendered results and the corresponding RGB-D inputs. Based on the unified framework, we introduce a joint optimization approach to fully exploit the inter-relationships between geometry, texture, and camera pose, and describe an adaptive interleaving strategy to improve optimization stability and efficiency. Using differentiable rendering, an image-level adversarial loss is applied to further improve the 3D model, making it more photorealistic. 
Experiments on synthetic and real data using quantitative and qualitative evaluation demonstrated the superiority of our approach in recovering both fine-scale geometry and high-fidelity texture. 
Code is available at \url{https://adjointopti.github.io/adjoin.github.io/}.
\end{abstract}

}

\maketitle

\IEEEdisplaynontitleabstractindextext
\IEEEpeerreviewmaketitle



\IEEEraisesectionheading{\section{Introduction}\label{sec:introduction}}

\begin{figure*}[!t]
\centering
\includegraphics[width=0.99\linewidth]{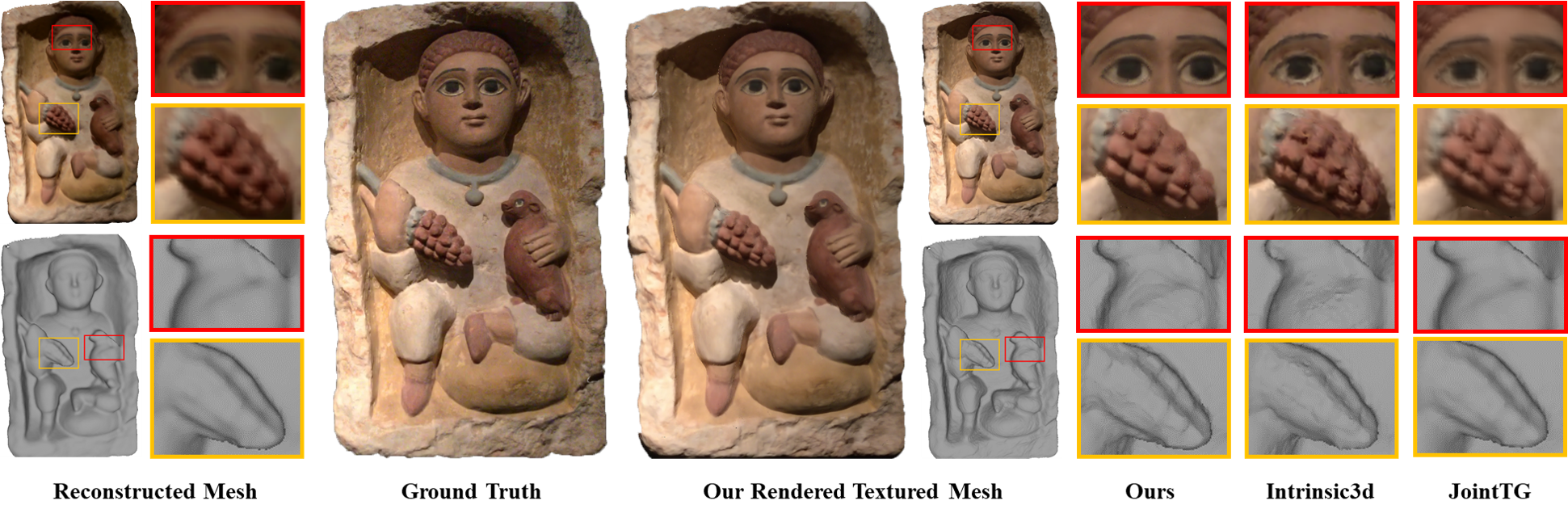}
\caption{A 3D mesh model reconstructed from RGB-D frames by KinectFusion lacks geometric and texture details. Compared with the state-of-the-art optimization schemes \cite{maier2017intrinsic3d,fu2020joint}, our method can obtain visually realistic textures and high-quality geometric models.}
\label{fig:teaser}
\end{figure*}

\IEEEPARstart{R}{econstructing} real-world 3D objects and scenes with high-fidelity texture and geometry has been a long-standing, important problem, since it has broad application prospects in various fields such as VR/AR, animations, and video games \cite{huang20173dlite, andersen2019ar,liu2019high}. With the emergence and wide availability of hand-held RGB-D cameras, it is now very convenient to reconstruct both the 3D geometry and the texture of an object. However, the reconstruction results are still far from satisfactory, as shown in Figure \ref{fig:teaser}. 
There are several mixed reasons for the production of such an inferior 3D textured model \cite{fu2020joint, bi2017patch}.
1) The depth noise produced by RGB-D cameras leads to imperfect 3D geometry. 2) The estimation errors of camera poses accumulate and finally lead to camera drifting. 3) Misalignment occur between the depth and color frames. 4) Due to the above errors, the texture mapping process, which naïvely projects multi-view images into another view-independent texture map, will inevitably produce blurring and ghosting artifacts.

In order to alleviate these problems, researchers have developed various algorithms to optimize the initial imperfect 3D model, and generate a higher-quality one by improving the texture quality \cite{bi2017patch,zhou2014color,li2018fast}, adjusting camera poses \cite{huang20173dlite,zhou2014color}, or refining mesh vertices \cite{wu2014real,zollhofer2015shading}. However, these methods, which focus on single-component optimization, subjectively break the inter-relationship between geometry, texture, and camera pose. Therefore, they have limited capabilities to compensate for reconstruction errors derived from the aforementioned mixed reasons. 
Maier \emph{et al.} \cite{maier2017intrinsic3d} first proposed an Intrinsic3D method to jointly optimize mesh, texture, camera pose, and scene lighting based on shape from shading (SFS). However, due to the SFS-based method's inherent limitations, Intrinsic3D is extremely slow and tends to suffer from the texture-copy problem (i.e., illogical geometric deformation caused by texture, shown in Figure~\ref{subfig_texcopy}). JointGT \cite{fu2020joint} effectively solves these problems by avoiding SFS in the optimization. Still, its complicated framework, which involves multiple optimization schemes and objective functions for geometry, texture, and camera pose, makes it less scalable and robust, especially when reconstruction errors are large. 

In this paper, we propose a unified framework for joint optimization of texture, geometry, and camera pose. This work was inspired by the recent successes of differentiable renderer \cite{li2018differentiable, kato2018neural, liu2019soft}. With it, the 3D model to be optimized can be rendered into multi-view images. Image-level objectives comparing the consistency between the rendered image and its corresponding RGB-D input can then be calculated, and back-propagated through the differentiable renderer to update the camera parameters, vertex positions, and texture colors of the 3D model, either separately or simultaneously. This unified framework has three benefits. First, with the unified framework, we jointly optimize the geometry, texture, and camera pose to fully exploit the inner relationships between different components and let them mutually improve each other. Second, our unified objectives are simple but effective. Unlike previous methods, which have to define specialized objectives defined on vertex color, vertex position, depth, and camera parameters, we use general image-level losses to supervise the optimization of geometry, texture, and camera pose. Third, thanks to the end-to-end framework with differentiable rendering, more advanced image-level objectives, such as perceptual loss \cite{johnson2016perceptual} and adversarial loss \cite{goodfellow2014generative} are introduced in our optimization, which greatly improves the rendering photorealism of 3D models optimized using our method. 

Considering the optimization stability and convergence rate, in our joint optimization framework we do not simply update the geometry, texture, and camera pose altogether in each iteration. Instead, we introduce an adaptive optimization strategy to smartly interleave the update of geometry, texture, and camera pose, based on the convergence of objectives. We performed quantitative and qualitative evaluations on our proposed joint optimization framework on various datasets. The experimental results demonstrated the value of our method in recovering both fine-scale geometry and high-fidelity texture, as shown in Figure~\ref{fig:teaser}.

To summarize, our contributions are three-fold:
\begin{itemize}
	\item We propose the first joint optimization framework for RGB-D reconstruction based on differentiable rendering. It unifies optimization schemes and objectives for geometry, texture, and camera pose; allows joint optimizations of different components to mutual benefit; and supports the use of adversarial loss to increase the photorealism of the reconstructed model.
	
	\item  We introduce a joint optimization strategy to adaptively interleave the update of geometry, texture, and camera pose, which leads to faster and more stable convergence. 
	
	\item  The experimental results show that our method performs considerably better than state-of-the-art methods, using either separate or joint optimization.
	
\end{itemize}

\begin{figure*}[!t]
\centering
\includegraphics[width=0.99\linewidth]{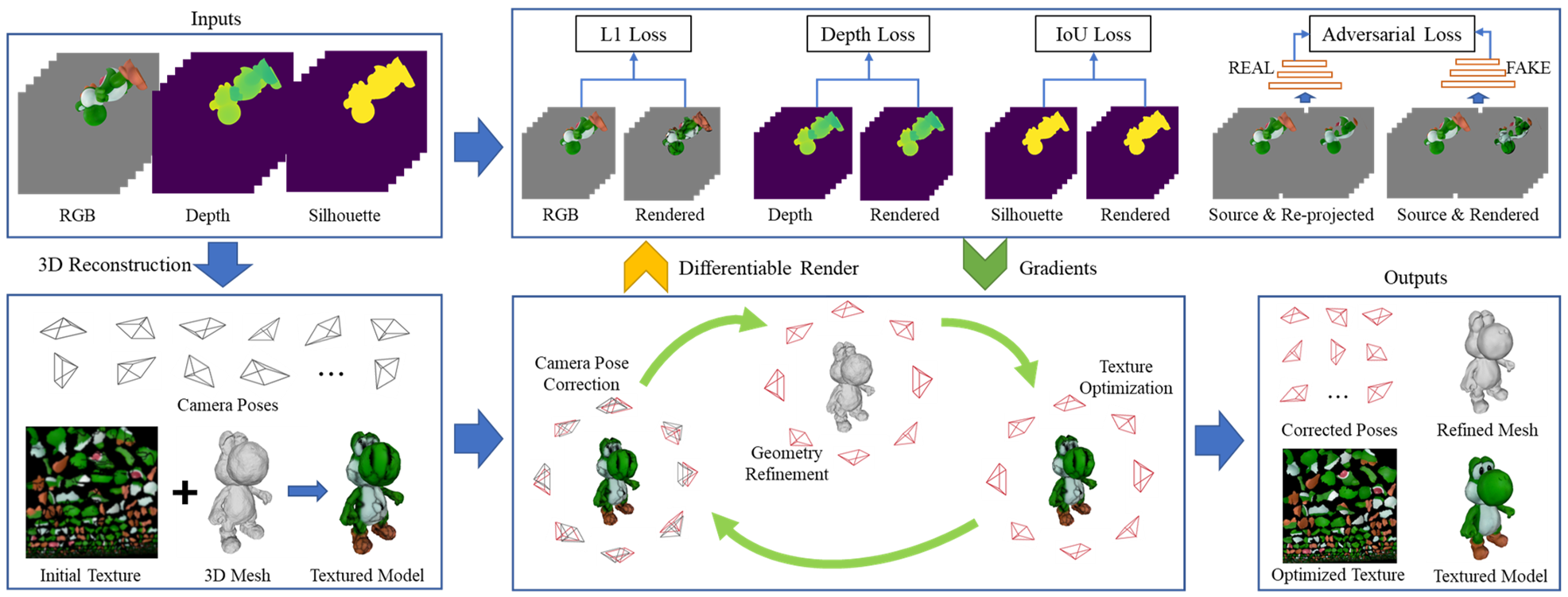}
\caption{Overview of our adaptive joint optimization method. The inputs include a set of RGB-D frames, estimated camera poses, and an initial imperfect 3D textured model produced by an existing reconstructing method. A differentiable renderer is adopted to produce rendered frames of the inputted camera poses. Then, a series of losses are employed to measure the consistency of the rendered and inputted frames, and the gradients are back-propagated to adaptively update camera pose, geometry, and texture. Finally, the corrected camera poses, refined geometry, and optimized texture are outputted.}
\label{fig:overview}
\end{figure*}


\section{Related Work}
\subsection{Geometry and Texture Optimization}
In this section, we provide a brief overview of texture and geometry optimization for 3D reconstruction. The first research line focuses on \textit{geometry refinement}. Zollh{\"o}fer \emph{et al.} \cite{zollhoefer2015shading} optimized the geometry of reconstructed 3D model encoded in a truncated signed distance field (SDF). Choe \emph{et al.} \cite{choe2017refining} exploited shading cues captured from infrared cameras to improve the quality of a 3D mesh. Romanoni \emph{et al.} \cite{romanoni2017multi} refined the surface geometry by optimizing composite energy in a variational manner, meanwhile updating the semantic labels based on Markov Random Field (MRF). Jiang \emph{et al.} \cite{jiang20183d} combined both facial priors and SFS to enhance the fine geometric details of a portrait model. Schmitt \emph{et al.} \cite{schmitt2020joint} recovered geometry details via jointly correcting camera pose and estimated material properties.

Another research line is about \textit{texture optimization}. Zhou \emph{et al.} \cite{zhou2014color} employed local warping for texture images to rectify complex distortions of geometry. Bi \emph{et al.} \cite{bi2017patch} used patch-based synthesis to correct the misalignment of multi-view reference images. Fu \emph{et al.} \cite{fu2018texture} proposed a  global-to-local non-rigid optimization method to achieve better texture mapping results. Li \emph{et al.} developed a fast texture mapping scheme to reduce misalignment at texture  boundaries. Lee \emph{et al.} adopted a texture-fution method with an SDF voxel grid to optimize the texture of a real-time scanning model.
Huang \emph{et al.} \cite{huang2020adversarial} designed a misalignment-tolerant metric based on generalized adversarial networks (GANs) to produce photorealistic textures. 

Different from the above-mentioned works, some recent research has attempted to \textit{optimize texture and geometry jointly}. Wang \emph{et al.} \cite{wang2018plane} used planar primitives to partition a model and jointly optimized plane parameters, camera poses, texture and geometry using photometric consistency and planar constraints. However, this method relies on plane priors, and is not suitable for complex nonplanar objects. Maier \emph{et al.} \cite{maier2017intrinsic3d} proposed an SFS-based method to simultaneously optimize texture, camera pose, geometry, and lighting, but SFS is an ill-posed problem with potentially ambiguous solutions. Such ambiguity will lead to inferior results with texture-copy artifacts. Fu \emph{et al.} \cite{fu2020joint} suggested directly optimizing the 3D models and avoiding the SFS process, which successfully alleviates the problem of texture-coping. However, they used different optimization schemes and objectives for texture, geometry, and camera pose optimizations, which makes the system complicated and less robust, especially to large reconstruction errors. Unlike Fu \emph{et al.} \cite{fu2020joint}, we propose a unified framework to jointly optimize texture, geometry, and camera poses via differentiable rendering. Experiments showed that our unified framework is more robust to different levels of reconstruction errors and achieves better results than existing methods. 

\subsection{Differentiable Rendering}
Triangle meshes, a well-established representation of a surface, have been used in almost every graphics pipeline due to their efficiency and flexibility in terms of vertex transformations and texturing. However, traditional graphics engines do not produce usable gradients for optimization purposes. To solve this problem, some early work \cite{kato2018neural,loper2014opendr} approximated gradients on mesh vertices, and recent methods \cite{chen2019learning,liu2019soft} propose fully-differentiable formulations. The property of differentiable renderer bridges the connection between a traditional optimization pipeline and advanced deep learning techniques. We adopt a soft-rasterizer renderer \cite{liu2019soft} in this work, which allows us to optimize the parameters of a 3D model, including texture, geometry, and camera poses, by enforcing weakly supervised losses between the rendered images and the corresponding RGB-D inputs.

\subsection{Generative Adversarial Networks}
GANs \cite{goodfellow2014generative} have achieved impressive progress in areas such as super-resolution \cite{Wang_2018_ECCV_Workshops}, image restoration \cite{Wan_2020_CVPR}, image translation \cite{isola2017image}, and 3D topological representation \cite{li2021sg}. The main principle of a GAN is to run a zero-sum game between a generator and a discriminator in an adversarial manner. Unlike previous photometric approaches such as L1 or L2, which will lead to over-smooth results, adversarial loss can ensure high-quality image generation with high-frequency details. However, it is difficult to include adversarial loss in a traditional optimization pipeline \cite{fu2020joint, li2018fast} due to the unstable process caused by entangled objectives. Huang \emph{et al.} \cite{huang2020adversarial} first propose to combine texture optimization and GANs to solve the misalignment problem through view transformation-based warping. In this work, we combine GANs with differentiable rendering to achieve effective joint texture and geometry optimization for 3D reconstruction.


\section{Methods}
Our method aims to produce both fine-scale geometric detail and high-fidelity texture for 3D model reconstructed from scanned RGB-D images. To this end, we propose a joint optimization framework to refine inaccurate camera poses, rough geometry surfaces, and unclear texture.

\subsection{Problem Setting}
The inputs of our method include a set of RGB-D images scanned by a consumer range camera, corresponding estimated camera poses and object silhouettes for each frame, and an initial imperfect 3D model reconstructed using an existing reconstructing method like BundleFusion \cite{dai2017bundlefusion}, KinectFution \cite{newcombe2011kinectfusion}, or COLMAP \cite{schonberger2016structure}. 
Let ${M}$ indicate the reconstructed 3D mesh, and ${T}$ represent the texture map to be optimized. 
$\boldsymbol{I}_{A}$, $\boldsymbol{D}_{A}$, and $\boldsymbol{S}_{A}$ respectively denote the color image, depth map, and silhouette map under view ${A}$, and its corresponding camera pose, indicated as $\boldsymbol{P}_{A}$, is composed of a rotation matrix and a translation vector, which can transform each vertex, $\boldsymbol{v}_{i}$, of a mesh, ${M}$, from the world coordinate system to the camera coordinate system by $\left[\widetilde{\boldsymbol{v}}_{i}\right]_{\text {world}}=\boldsymbol{P}_{A}\left[\widetilde{\boldsymbol{v}}_{i}\right]_{\text {camera}}$, 
where $\left[\widetilde{\boldsymbol{v}}_{i}\right]_{X}$ indicates the homogeneous coordinate of the vertex $\boldsymbol{v}_{i}$ in the $X$ coordinate system. Now the problem is how to produce the optimized 3D reconstruction, based on known information.

\subsection{Overview}
Considering that 3D reconstruction errors come from mixed factors regarding both texture and geometry, we jointly optimize the vertex position, camera pose, and texture map with a unified framework based on differentiable rendering. As depicted in Sec.~\ref{sec: framework}, the 3D mesh, ${M}$, with texture, ${T}$, is rendered with different camera poses, $\boldsymbol{P}$. Then a series of image-level losses measuring the differences between the rendered frames and the RGB-D inputs are computed, to update ${M}$, ${T}$, and $\boldsymbol{P}$. To achieve fast and stable convergence, we do not naïvely update ${M}$, ${T}$, and $\boldsymbol{P}$ together in every iteration. Instead, we designed a strategy to adaptively interleave the optimization of ${M}$, ${T}$, and $\boldsymbol{P}$ according to the convergence rate in Sec.~\ref{sec:strategy}. The overall pipeline is shown in Figure~\ref{fig:overview}.

\subsection{Unified Optimization Framework}\label{sec: framework}
To enable the effective and efficient joint optimization of the different components of a 3D model, a unified framework supporting the optimization of texture, geometry, and camera pose is essential. However, this task is not easy, since texture, geometry, and camera pose are in different domains. Specialized schemes and objectives have often been designed for the optimization of each component in previous work. 
Inspired by the fact that the rendering process is a function combining texture, geometry, and camera pose, we propose a novel unified optimization framework based on differentiable rendering. Next, we describe the details of this framework including the multi-view rendering process and common objectives.

\subsubsection{Multi-view Rendering}
During optimization, a 3D mesh ${M}^{t}$ with texture map ${T}^{t}$ and an associated camera pose $\boldsymbol{P}^{t}_{A}$ are fed to the differentiable renderer to generate a color image $\boldsymbol{I}_{A}^{DR,t}=DR\left(M^{t}, T^{t} \mid \boldsymbol{P}_A^{t}\right)$, depth image $\boldsymbol{D}_{A}^{DR,t}$ and silhouette image $\boldsymbol{S}_{A}^{DR,t}$, where ${DR}$ indicates the differentiable rendering operation. 
Instead of directly approximating $\boldsymbol{I}_{A}^{DR,t}$ with its ground truth $\boldsymbol{I}_A$ in view $A$, we generate more ground-truth references for comparison, by re-projecting RGB-D inputs under other auxiliary views to the target view $A$. This approach can help to compensate for errors in a single view and is inspired by \cite{huang2020adversarial}. Specifically, we sequentially select a color image $\boldsymbol{I}_{A}$ with associated camera pose $\boldsymbol{P}_{A}$ from all the candidate RGB-D inputs as the target view, and randomly select another image $\boldsymbol{I}_{B}$ with camera pose $\boldsymbol{P}_{B}$ from adjacent views of $\boldsymbol{I}_{A}$ as the auxiliary view. For each pixel $\boldsymbol{q}_{B}$ in the auxiliary image $\boldsymbol{I}_{B}$, we can compute its re-projected pixel $\boldsymbol{q}_{B\rightarrow A}$ via a simple spatial transformation:

\begin{equation}
\boldsymbol{q}_{B\rightarrow A}=\boldsymbol{K} \boldsymbol{P}_{A} \boldsymbol{P}_{B}^{-1} \left(d_{B}\boldsymbol{K}^{-1}\boldsymbol{q}_{B}\right),
\label{equ:reproj}
\end{equation}
where $d_{B}\boldsymbol{K}^{-1}\boldsymbol{q}_{B}$ represents the associated vertex position of view $B$ in camera space, which is produced by using the known depth value $d_{B}$ and intrinsic matrix $\boldsymbol{K}$ of the camera. Then, the two projection transformations---camera to world and world to camera---are employed to find the corresponding vertex position of the camera coordinate of expected view $A$. 
According to the transformation of a single pixel in Eq.~\ref{equ:reproj}, we can re-project the RGB image under auxiliary view ${B}$ into the target view $A$ by: $\boldsymbol{I}_{B \rightarrow A}=\operatorname{RP}\left(\boldsymbol{I}_{B}\rightarrow\boldsymbol{I}_{A}\right)$, where $\operatorname{RP}$ represents the re-projection process.

To avoid the extreme case where the re-projected image has few visible pixels, the auxiliary view $B$ will be selected within a maximum $15^{\circ}$ deviation of the target view $A$.
Practically, the neighboring views of $A$, denoted as $J_{A}$, are determined using the following metric:
\begin{equation}
J_{A}=\left\{\boldsymbol{I}_{B} \in \Psi_{\text {color}}: \angle\left(\boldsymbol{R}_{A}, \boldsymbol{R}_{B}\right) \leq 15^{\circ}\right\}
\label{eq:multiview}
\end{equation}
where $\Psi_{\text {color}}$ is the color image set, and $\angle\left(\boldsymbol{R}_{A}, \boldsymbol{R}_{B}\right)$ denotes the angle between the rotation matrix $\boldsymbol{R}_{A}$ and $\boldsymbol{R}_{B}$ in view $A$ and $B$.

\subsubsection{Common Objectives}
In every iteration $t$, the rendered color image $\boldsymbol{I}_{A}^{DR,t}$, depth image $\boldsymbol{D}_{A}^{DR,t}$ and silhouette $\boldsymbol{S}_{A}^{DR,t}$ in view $A$ are compared with the re-projected image $\boldsymbol{I}_{B \rightarrow A}$, scanned depth map $\boldsymbol{D}_{A}$ and object mask $\boldsymbol{S}_{A}$, to achieve a common objective of reconstructing RGB-D inputs. The most fundamental loss is image-level L1, which is employed to guide the optimization process based on the re-projecting transformation, to enhance the color consistency between the rendered image and the ground truth, defined as follows:
\begin{equation}
L_{RGB}=\left\| \boldsymbol{I}_{B \rightarrow A}-\boldsymbol{I}_{A}^{DR,t}\right\|_{1}.
\end{equation}

The depth loss $L_{depth}$ is calculated by measuring the L1 distance between the scanned and rendered depth images:

\begin{equation}
L_{depth}=\left\| \boldsymbol{D}_{A}-\boldsymbol{D}_{A}^{DR,t}\right\|_{1},
\end{equation}
which is effective to enhance the geometric consistency. 

We employ an Intersection-over-Union (IoU) loss to measure the silhouette consistency between mesh model and object:

\begin{equation}
L_{IoU}=1-\frac{\left\|\boldsymbol{S}_{A} \otimes \boldsymbol{S}_{A}^{DR,t}\right\|_{1}}{\left\|\boldsymbol{S}_{A} \oplus \boldsymbol{S}_{A}^{DR,t}-\boldsymbol{S}_{A} \otimes \boldsymbol{S}_{A}^{DR,t}\right\|_{1}},
\end{equation}
where $\otimes$ and $\oplus$ represent the element-wise product and the sum operator, respectively.

Finally, we take the weighted sum of the RGB loss, the depth loss, and the IoU loss as the common objective to optimize texture color $T^t$, mesh vertices $M^t$, and camera parameters $\boldsymbol{P}^t$:

\begin{equation}
L_{common}=\lambda_{C} L_{RGB} + \lambda_{D}L_{depth} + \lambda_{S}L_{IoU},
\label{eq:common}
\end{equation}
where $\lambda_{C}$, $\lambda_{D}$ and $\lambda_{S}$ are constant coefficients balancing between terms. We empirically set $\lambda_{C}=0.1$, $\lambda_{D}=1$ and $\lambda_{S}=1$ in all the experiments.
Note that in each epoch, this common objective (Eq.~\ref{eq:common}) under all views is averaged.

\subsection{Joint Optimization} \label{sec:strategy}
With our unified optimization framework, one naïve joint optimization strategy is to update the texture color $T^t$, mesh vertices $M^t$, and camera parameters $\boldsymbol{P}^t$ together in every single iteration $t$. However, we found this naïve strategy results in slow and unstable convergence. To solve this problem, we developed a joint optimization strategy to adaptively interleave updates of different components based on the common objective's convergence rate (Eq.~\ref{eq:common}). 
With this strategy, the updating iterations for each component will be dynamically adjusted in different stages, leading to more stable and efficient optimization. In this way, the optimization of different components will still be beneficial to each other across iterations.   

In addition to convergence acceleration, the interleaving strategy brings another benefit: it allows the addition of specialized objectives for a component in its own iteration. Specifically, we add Laplacian loss in the geometry iterations to constrain the mesh smoothness, while adding adversarial loss in the texture iterations to increase photorealism. Next, we introduce specific losses used in geometry and texture iterations, and then our adaptive interleaving strategy.

\subsubsection{Geometry Iteration with Laplacian Loss}
The initial geometry of an input 3D mesh usually suffers from noise introduced during the reconstruction procedure. In geometry iteration, vertex positions are adjusted to restore high-fidelity geometry by maximizing the consistency between the rendered results and the scanned RGB-D images. To further guarantee the local smoothness of the refined mesh, we add a Laplacian loss to the common objective (Eq.~\ref{eq:common}) to refine the geometry. Let $\boldsymbol{W}$ denote the uniform Laplacian matrix \cite{nealen2006laplacian} of size $m \times m$, where $m$ is the number of vertices. The Laplacian loss can be defined as an L2 norm of the Laplacian coordinates:
\begin{equation}
L_{Lap}=\left\| \boldsymbol{W}\boldsymbol{V}_{d} \right\|_{2},
\label{eq:lapla}
\end{equation}
where $\boldsymbol{V}_{d} = \left[\boldsymbol{v}_{1d},\boldsymbol{v}_{2d},\cdots,\boldsymbol{v}_{md}\right]^{T}$ represents the vertex matrix of mesh $M^t$, and $d\in {x, y, z}$ indicates the spatial coordinate axis.

\subsubsection{Texture Iteration with Adversarial Loss}
Using only the common objective (Eq.~\ref{eq:common}) defined by the L1 norms, the optimized texture will appear blurry and lose fine detail. To solve this problem, we introduce adversarial learning in the texture iteration in addition to $L_{common}$. By jointly training an image discriminator network $\mathcal{D}$ to distinguish 'real' and 'fake' images, we aim to produce a texture that is indistinguishable from re-projections of captured images from other views. Accordingly, the adversarial loss is defined as:

\begin{equation*}
\begin{aligned}
L_{adv}=\log \mathcal{D}\left(\boldsymbol{I}_{A}, \boldsymbol{I}_{B\rightarrow A}\right) +\log \left(1-\mathcal{D}\left(\boldsymbol{I}_{A}, \boldsymbol{I}_{A}^{DR,t}\right)\right)
\end{aligned}
\end{equation*}

During optimization, the discriminator $\mathcal{D}$ learns to recognize artifacts like seams, noise, or blurriness, that do not appear in real images and is tolerant to small misalignment between a rendered image $\boldsymbol{I}_{A}^{DR,t}$ and the real image $\boldsymbol{I}_{B\rightarrow A}$. Therefore, the texture optimized to fool the discriminator appears more realistic than that produced using only the common objective.

\begin{figure}[!t]
  \label{alg:strategy}
  \removelatexerror
  \begin{algorithm}[H]
    \caption{Pseudocode for Adaptive Iterative Strategy}
    \begin{algorithmic}[0]
      \renewcommand{\algorithmicrequire}{\textbf{Input:}}  
      \REQUIRE
        \STATE threshold $\delta=10^{-3}$, patience $\Omega=50$;
        \STATE common loss $L=\left[L^{(1)}, \cdots, L^{(t)}, \cdots\right]$.
      \renewcommand{\algorithmicrequire}{\textbf{Initialize:}} 
      \REQUIRE
        \STATE best value $\beta=L^{(1)}$, maximum steps $T_{max}$ (e.g., $10^{3}$)
      \renewcommand{\algorithmicrequire}{\textbf{Decision-Making Process:}} 
      \REQUIRE
        \STATE\textbf{Let} counter $n = 0$, $\text{strategy} = keep$
        \FOR{$t=0$ to $T_{max}$}
          \IF{$L^{(t)}< \beta*(1-\delta)$}
            \STATE $n=0$, $\beta=L^{(t)}$;
          \ELSE \STATE $n += 1$
          \ENDIF
          \WHILE{$n \geq \Omega$}
            \STATE $\text{strategy} = next$
          \ENDWHILE
        \ENDFOR
      \renewcommand{\algorithmicensure}{\textbf{Return:}}
      \ENSURE $\text{strategy}$
    \end{algorithmic}
  \end{algorithm}
\end{figure}

\subsubsection{Adaptive Interleaving Strategy}
In this strategy, we employ external iterations to achieve joint optimization and internal iterations to search for a temporary optimum based on current data for each stage. Specifically, in each external iteration, we first minimize $L_{common}$ to correct the camera poses $\boldsymbol{P}$ until a local optimum is reached, while fixing geometry $M$ and texture $T$. Then, the 3D mesh $M$ is refined by minimizing the sum of the common loss and additional Laplacian loss, with both $\boldsymbol{P}$ and $T$ fixed. Finally, we optimize the texture map $T$ based on the updated and fixed $\boldsymbol{P}$ and $M$, using adversarial learning. The external iteration cycle among camera pose, geometry, and texture is repeated three times. Interleaving the optimization of each component makes the convergence more stable.

The internal iteration numbers in each external iteration are not set manually. Instead, an adaptive strategy is adopted, to achieve a good trade-off between convergence speed and quality. The proposed adaptive strategy tested after each internal iteration is illustrated in Algorithm \ref{alg:strategy}. It checks the convergence rate of the common objective $L_{common}$, and returns a flag $next$ to terminate internal iteration, or otherwise continue.


\begin{figure*}[!t]
\centering
\includegraphics[width=1\linewidth]{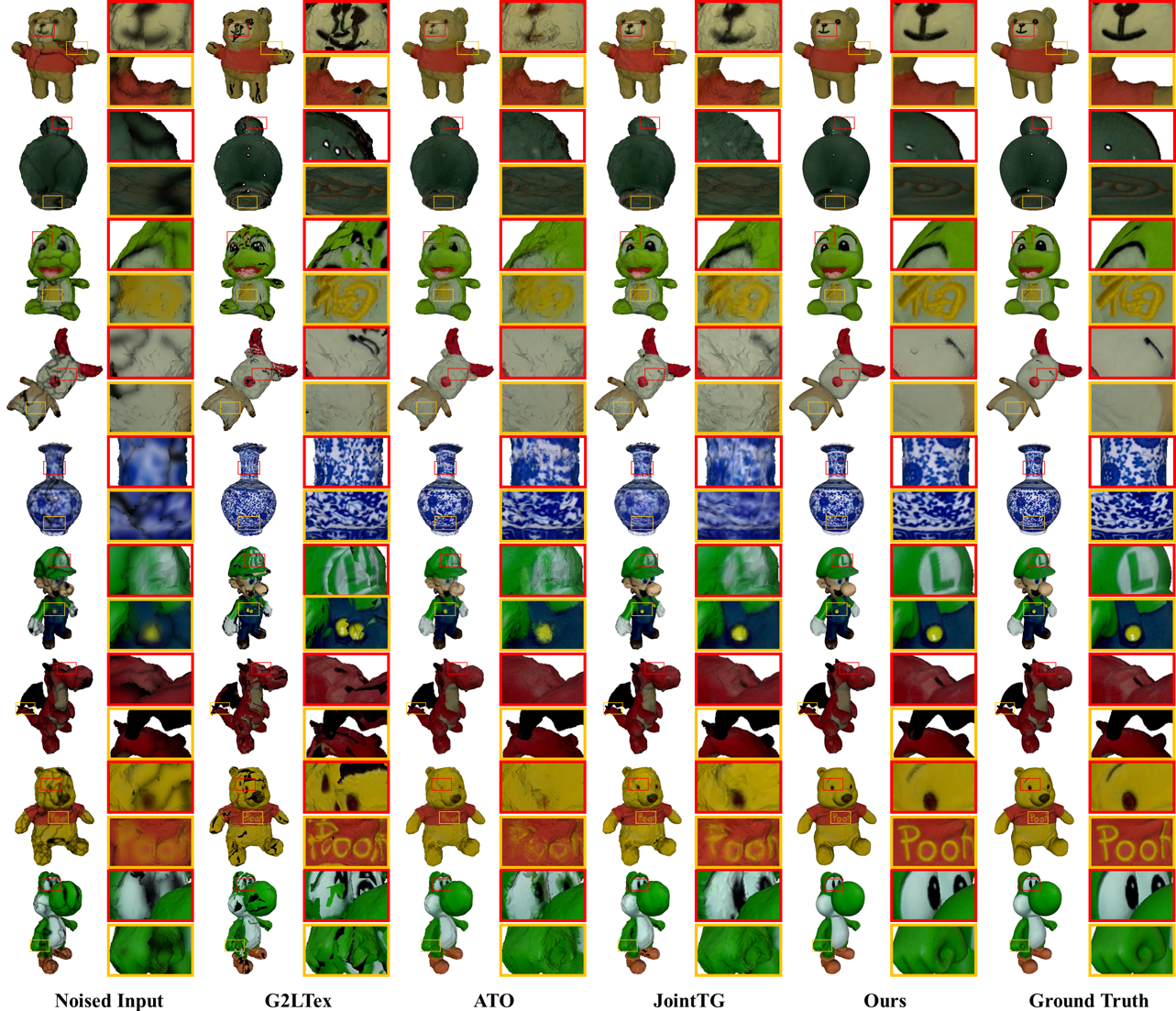}
\caption{Visual comparison of optimization results produced by different methods on synthetic data.}
\label{fig:ex1_1}
\end{figure*}

\section{Experiments}
\label{sec:ex}
We first compared our method with state-of-the-art methods on a synthetic dataset by adding noise to high-quality laser scanning models. Then, we evaluated our method on public RGB-D datasets to demonstrate the effectiveness of our method on real 3D reconstructions. We used ablation studies to validate the performance of each major component of our pipeline. An extension study using our method for the optimization of RGB data is described in Sec.~\ref{sec:exten}. Finally, we performed a user study to qualitatively compare the experimental results of the different methods on synthetic and real data.
All of the experiments were conducted on a PC with Intel Core i7-9700 3.00GHz and GeForce RTX2080Ti 12GB, using author-released codes for compared methods \cite{maier2017intrinsic3d, fu2018texture, huang2020adversarial, fu2020joint}.

\begin{figure}[!t]
\centering
\subfloat[PSNR $\uparrow$]{
\begin{minipage}[t]{0.49\linewidth}
\centering
\includegraphics[width=1.6in]{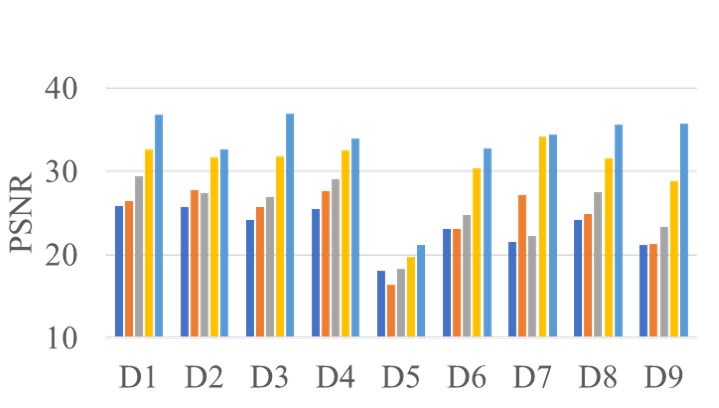}
\end{minipage}%
}%
\subfloat[SSIM $\uparrow$]{
\begin{minipage}[t]{0.49\linewidth}
\centering
\includegraphics[width=1.6in]{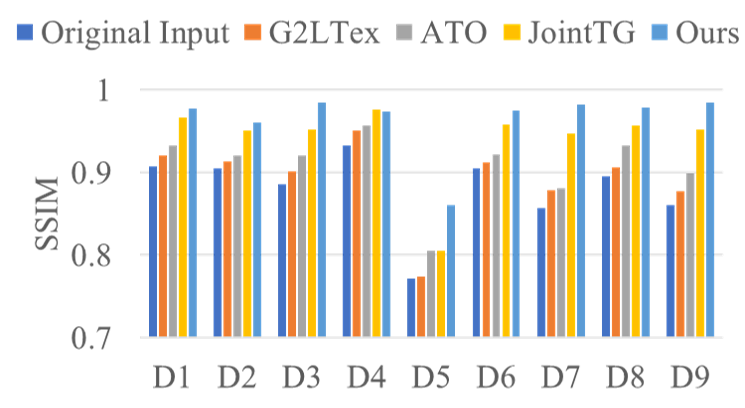}
\end{minipage}%
}%
\quad
\subfloat[Perceptual Loss $\downarrow$]{
\begin{minipage}[t]{0.49\linewidth}
\centering
\includegraphics[width=1.6in]{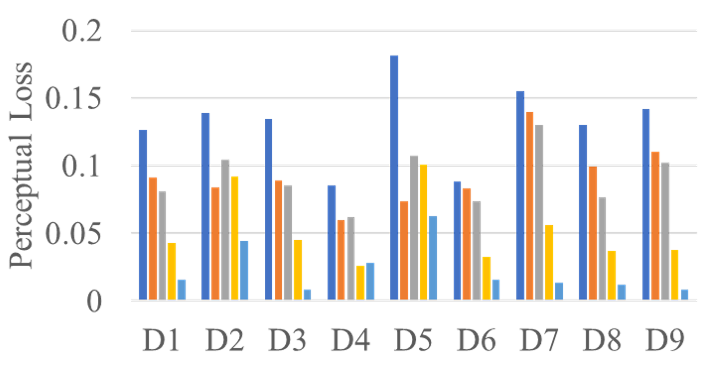}
\end{minipage}
}%
\subfloat[Hausdorff Distance $\downarrow$]{
\begin{minipage}[t]{0.49\linewidth}
\centering
\includegraphics[width=1.6in]{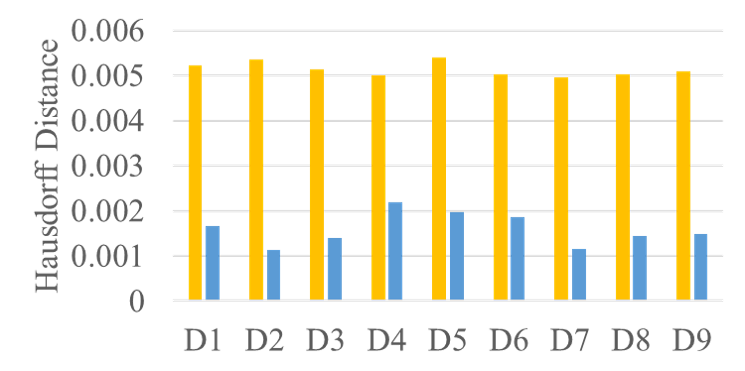}
\end{minipage}
}%
\centering
\caption{The PSNR, SSIM, and Perceptual loss of results optimized by the state-of-the-art methods and ours is displayed in (a), (b), and (c). Due to G2LTex \cite{fu2018texture} and ATO \cite{huang2020adversarial} only focus on texture optimization without geometry refinement, we only report the Hausdorff distance of the optimized mesh produced by JointTG \cite{fu2020joint} and the proposed method in (d).}
\label{fig:ex1_2}
\end{figure}

\begin{figure}[!t]
\centering
\subfloat[Camera Pose Errors]{
\begin{minipage}[t]{0.49\linewidth}
\centering
\includegraphics[width=1.6in]{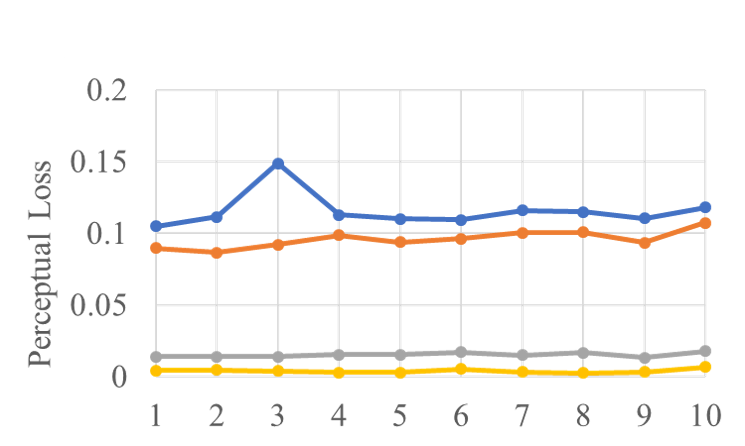}
\end{minipage}%
}%
\subfloat[Geometry Errors]{
\begin{minipage}[t]{0.49\linewidth}
\centering
\includegraphics[width=1.6in]{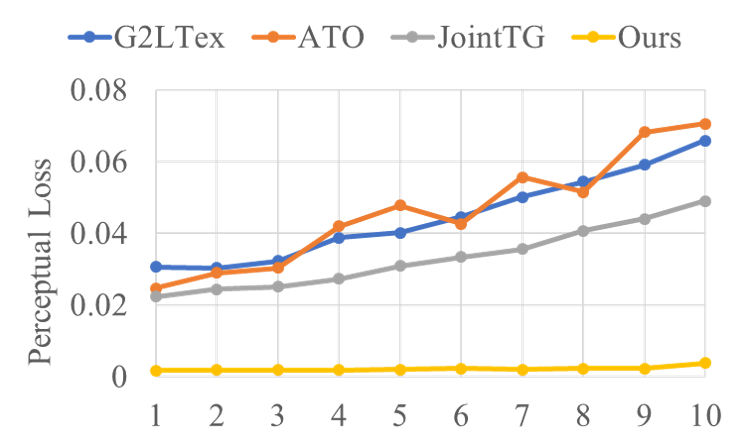}
\end{minipage}%
}%
\centering
\caption{Robustness study: comparison of optimization results produced by different methods with increasing level of noises on synthetic data. (a) indicates the Perceptual loss curve changing as the degree $n=0.25N$ of camera-pose noise increases from $N$=1 to 10. (b) indicates the Perceptual loss curve changing as the degree $n=0.25N$ of geometry noise increases from $N$=1 to 10.}
\label{fig:ex1_noised}
\end{figure}

\subsection{Evaluation on Synthetic Data}
\label{sec:syn}
To quantitatively evaluate the performance of our method, we first tested it on a synthetic dataset for which both geometry and texture ground truth are known. Three representative baselines were compared with our method: G2LTex \cite{fu2018texture}, in which only texture is optimized via a texture mapping strategy; Adversarial Texture Optimization (ATO) \cite{huang2020adversarial}, in which only texture is optimized with adversarial loss; and JointTG \cite{fu2020joint}, in which camera pose, geometry, and texture are jointly optimized. Here, we ignored Intrinsic3d \cite{maier2017intrinsic3d}, since its model refinement is based on a signed distance field (SDF) volume instead of a given noisy textured mesh, although this method was used in experiments with real data.

\subsubsection{Synthetic Dataset}
We collected nine high-quality 3D models (D1-D9) reconstructed by high-precision laser scanning to serve as the ground truth. We uniformly sampled 40 views on a unit sphere, and rendered the textured model into these views, to simulate the RGB-D images scanned by a consumer range camera. For each ground truth model, random noise was added to its geometry, texture, and camera poses, to synthesize an imperfect input model. 

Specifically, we added uniformly distributed noise ranging from $\left[-e_{t}, e_{t}\right]$ to the translation vector, and noise of Euler angle ranging from $\left[-e_{r}, e_{r}\right]$ to the rotation matrix, to simulate inaccurate camera poses. For geometric errors, we added random disturbances to the mesh vertices, and applied three steps of Laplacian smoothing to these random values, in which the disturbances satisfy a uniform distribution ranging from $\left[-e_{g}, e_{g}\right]$. In our experiment, we set ${e}_{t}=0.01 \times 1.5^{n}$, $e_{r}=5$, and $e_{g}=0.03\times1.5^{n}$, where $n$ is a measure of the degree of noise. We set $n=1.5$ in the general comparative experiment and changed $n$ from $0.25$ to $2.5$ at intervals of $0.25$ in the robustness study of the optimization methods. For texture errors, we intentionally misaligned and blurred the original texture, and randomly added irregular masks to simulate misalignment, blurriness, and seams on the textures. Some synthesized data are shown in the first column of Figure~\ref{fig:ex1_1}.

\begin{table*}[!t]
\renewcommand{\arraystretch}{1.3}
\caption{Quantitative comparison on the state-of-the-art optimization methods and ours for real datasets.}
\label{tab:evluation_ex2}
	\centering
    \scalebox{0.9}[1]{
		\begin{tabular}{lcccccccccccc}
            \hline
			{Datasets} & \multicolumn{3}{c}{Dolls (9 models)} & \multicolumn{3}{c}{Intrinsic3d (5 models)} & \multicolumn{3}{c}{Scene3d (12 models)} & \multicolumn{3}{c}{Chairs (35 models)}\\
			{} & PSNR$\uparrow$ & SSIM$\uparrow$ & Perceptual$\downarrow$ & PSNR$\uparrow$  & SSIM$\uparrow$ & Perceptual$\downarrow$ & PSNR$\uparrow$ & SSIM$\uparrow$ & Perceptual$\downarrow$ & PSNR$\uparrow$ & SSIM$\uparrow$ & Perceptual$\downarrow$\\
			\hline
			G2LTex & 22.758 & 0.920 & 0.070 & 21.684 & 0.649 & 0.276 & 17.637 & 0.583 & 0.287 & 23.316 & 0.791 & 0.167\\
			ATO & 23.734 & 0.941 & 0.058 & 25.463 & 0.770 & 0.269 & 21.235 & 0.680 & 0.250 & 25.011 & 0.819 & 0.148\\
			Intrinsic3d & 23.171 & 0.929 & 0.079 & 24.552 & 0.727 & 0.253 & 17.390 & 0.576 & 0.361 & 22.523 & 0.781 & 0.209\\
			JointTG & 23.354 & 0.930 & 0.079 & 22.262 & 0.697 & 0.327 & 19.756 & 0.640 & 0.325 & 24.113 & 0.820 & 0.214\\
			Ours & \textbf{25.185} & \textbf{0.949} & \textbf{0.042} & \textbf{26.716} & \textbf{0.793} & \textbf{0.251} & \textbf{21.462} & \textbf{0.696} & \textbf{0.247} & \textbf{26.001} & \textbf{0.831} & \textbf{0.135}\\
            \hline
		\end{tabular}
		}
	\normalsize
\end{table*}


\subsubsection{Evaluation Metric}
In order to measure the quality of the optimized texture and geometry compared to the ground truth of the synthetic dataset, we employed several evaluation metrics: peak-signal-to-noise ratio (PSNR), structural similarity index measure (SSIM), and Perceptual loss \cite{zhang2018unreasonable}.
Considering that rendered results integrate the camera poses, geometry, and texture of a 3D model, we calculated these metrics between the rendered and source images on the same viewpoint to comprehensively evaluate the overall quality of an optimized model.
To illustrate the performance of the optimization algorithms with respect to geometry details, we adopted the average Hausdorff distance to measure the difference between the optimized and ground truth meshes. Since G2LTex \cite{fu2018texture} and ATO \cite{huang2020adversarial} will not optimize mesh, the Hausdorff distance metric was only used to evaluate JointTG \cite{fu2020joint} and our method.

\subsubsection{Experimental Results}
We report the quantitative evaluation results in Figure \ref{fig:ex1_2}. Our method consistently outperformed all other baselines by a large margin on all metrics. This is also clearly reflected in Figure \ref{fig:ex1_1}, in which we show visual comparisons of the results optimized by state-of-the-art methods and ours. 
G2LTex \cite{fu2018texture} is a texture mapping method which does not consider camera-pose and geometry errors, divides the model surface into several regions, and maps the texture of one specific image to each region. This makes it locally sharper than other methods that blend the color information of all views, but easily causes texture misalignment artifacts (black seams) when camera pose noise exists.
Although ATO \cite{huang2020adversarial} also optimizes texture only, it is more robust to small amounts of camera-pose and geometry noise than G2LTex, because the adversarial loss, which we also adopted in our method, can tolerant small misalignment. Due to the joint optimization of camera pose, geometry, and texture, JointTG \cite{fu2020joint} performed better than either ATO or G2LTex. However, constrained by its complicated framework and different schemes used to optimize each component, it is less robust to camera and geometry noise than our method, especially when the errors are large. This point can be clearly observed in Figure \ref{fig:ex1_noised}. When the noise level increased in geometry or camera poses, our performance on perceptual loss was quite stable, while the performance of other methods, including JointTG \cite{fu2020joint}, dropped quickly. To summarize, compared to previous methods, our performance gain comes largely from the combined effects of three aspects: a unified framework, adversarial loss, and an adaptive interleaving strategy.

\begin{figure*}[!t]
\centering
\includegraphics[width=0.95\linewidth]{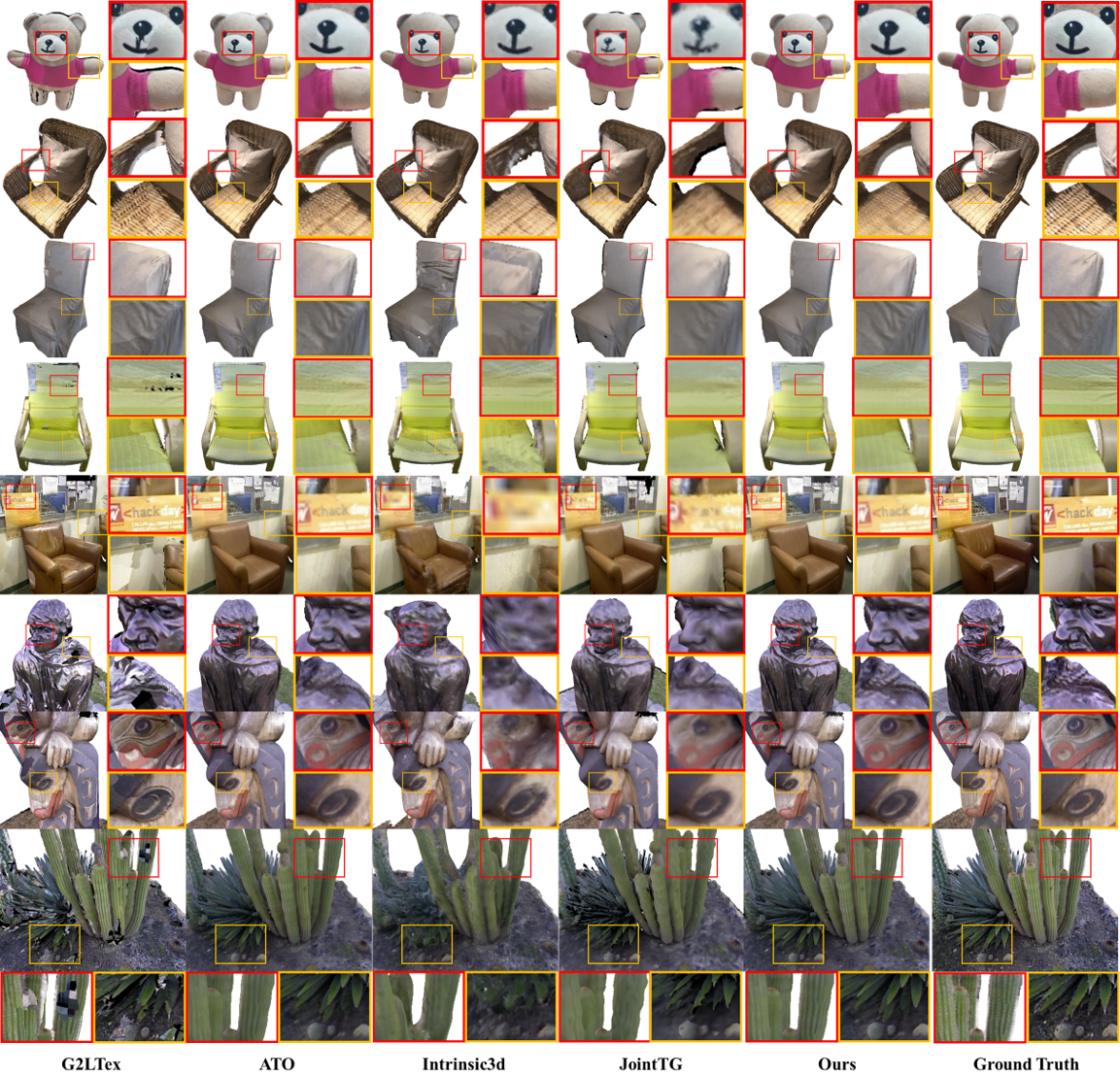}
\caption{Visual comparison of rendered results produced by different methods on real data.}
\label{fig:ex2_2}
\end{figure*}

\begin{figure*}[!t]
\centering
\includegraphics[width=0.95\linewidth]{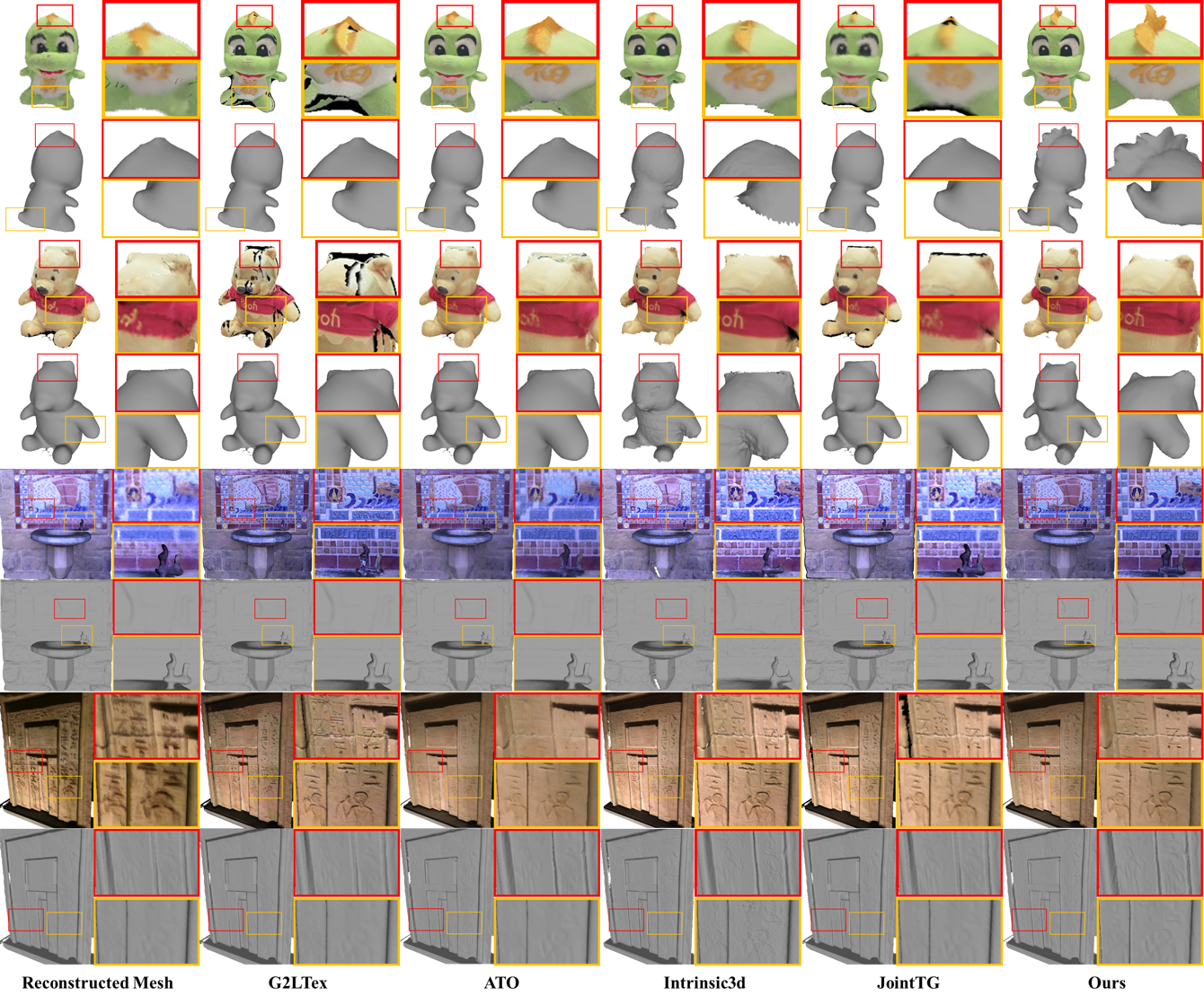}
\caption{Visual comparison of geometric details produced by different methods on real data.}
\label{fig:ex2_1}
\end{figure*}

\subsection{Evaluation on Real Data}

To demonstrate the performance of the proposed method on scanned objects and scenes in real scenarios, we compared our method to state-of-the-art baselines including two texture optimization methods (G2LTex \cite{fu2018texture} and ATO \cite{huang2020adversarial}) and two joint optimization methods (Intrisic3D \cite{maier2017intrinsic3d} and JointTG \cite{fu2020joint}) on real datasets. The datasets contain 61 models from our works, and that of Zhou \emph{et al.} \cite{zhou2014color}, Maier \emph{et al.} \cite{maier2017intrinsic3d}, and Huang \emph{et al.} \cite{huang2020adversarial}. Figure \ref{fig:ex2_2} and Figure \ref{fig:ex2_1} display the visual comparisons of the rendered images and the geometry details of the optimized 3D models. The corresponding quantitative evaluations using image quality metrics for different datasets are listed in the Table \ref{tab:evluation_ex2}. Here, the Hausdorff distance metric is not given, because no ground truth mesh was available for the real data. Our method performed consistently better than the other methods.

\begin{figure}[!t]
\centering
\subfloat[Geometric Deformation Experiment]{
\begin{minipage}[t]{0.99\linewidth}
\centering
\includegraphics[width=3.4in]{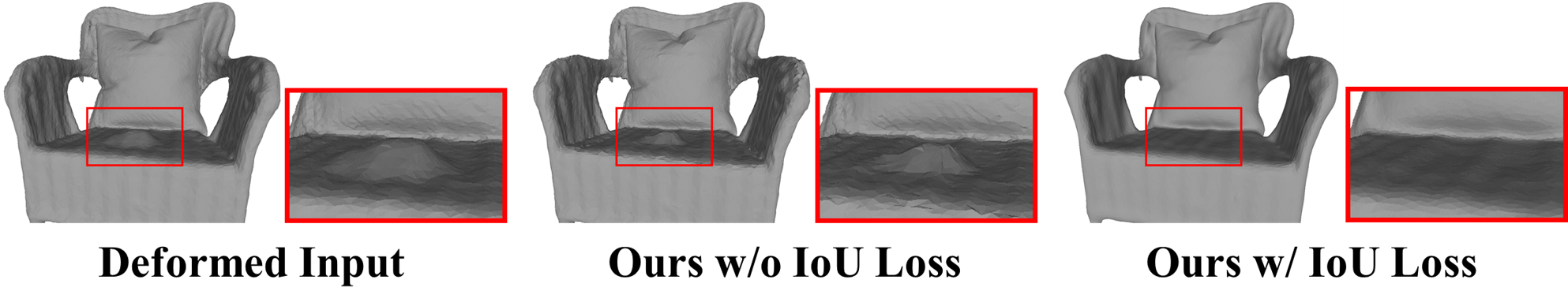}
\end{minipage}
\label{subfig_IoU}
}
\hfil  
\subfloat[Example of Texture-copy Artifacts]{
\begin{minipage}[t]{0.95\linewidth}
\centering
\includegraphics[width=3.4in]{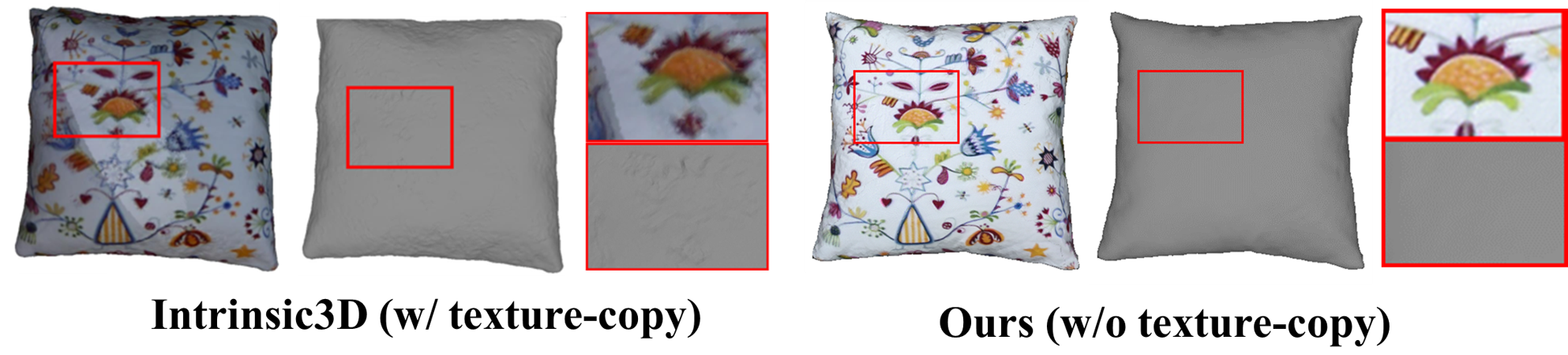}
\end{minipage}
\label{subfig_texcopy}
}
\hfil  
\subfloat[Optimization with RGB Data]{
\begin{minipage}[t]{0.95\linewidth}
\centering
\includegraphics[width=3.4in]{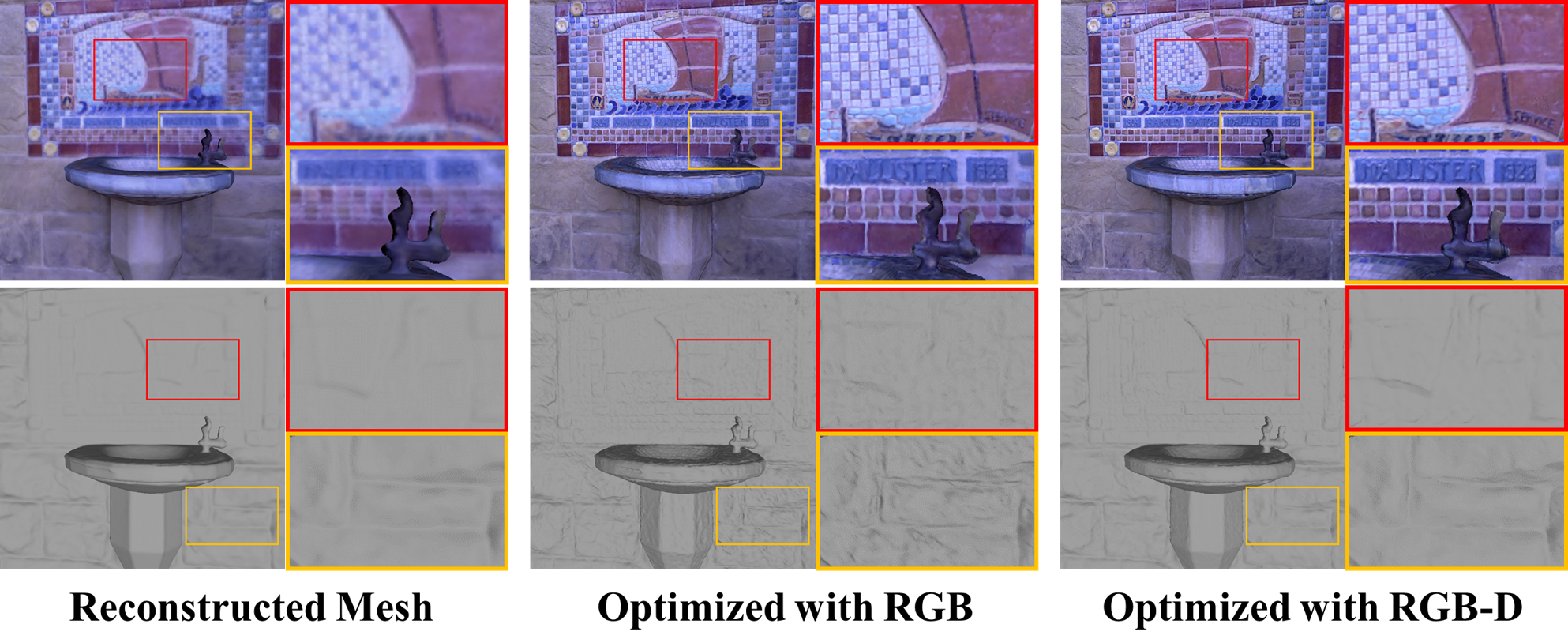}
\end{minipage}%
\label{subfig_RGB}
}
\centering
\caption{Examples of optimization on real data. (a) displays results of a geometric deformation experiment to illuminate the effectiveness of IoU loss. (b) gives example of texture-copy artifacts caused by Intrinsic3d method. (c) shows optimization results produced by our method based on RGB/RGB-D data.}
\label{fig:Iou_Depth}
\end{figure}

Qualitative comparisons also demonstrated the superior performance of our method in recovering both fine-scale geometry and high-fidelity texture. As shown in Figure \ref{fig:ex2_1}, G2LTex \cite{fu2018texture} based on texture mapping is sensitive to camera-poses and geometry noises, and finds it hard to obtain the correct texture when noise exists. ATO \cite{huang2020adversarial} obtains better texture by fusing texture information from multiple views and using adversarial loss. Still, it fails to recover geometric details, since only the texture is optimized. Intrinsic3d \cite{maier2017intrinsic3d} and JointTG \cite{fu2020joint} can refine the geometry details to some extent. However, Intrinsic3d, based on SFS, is sensitive to lighting and has texture-copy artifacts (shown in Figure \ref{subfig_texcopy}), especially when the views are sparse. The texture and mesh generated by JointTG are visually harmonious, but it fails to recover both geometric and texture details as accurately as our method. Therefore, its rendered results look blurrier. More importantly, no previous methods could recover imperfect geometry shapes from some rough initial models, because of their limited geometric deforming capabilities (see the first example of Figure \ref{fig:ex2_1}). Our method can handle geometric deformation problems, due to the IoU loss supported by our framework, as shown in Figure \ref{subfig_IoU}.

To further compare the performance of different methods on real data, we report the average running time on different datasets in Table \ref{tab:runningtime}. Since our method is implemented based on differentiable rendering, the running time of our method is linearly related to the model complexity and the number of scanned frames. As shown in Table \ref{tab:runningtime}, the compared methods were more sensitive to the model complexity and the number of frames than ours. Whereas our method took a similar time to other methods in the low-complexity models (\textit{Dolls} and \textit{Chairs}), our method was faster in the high-complexity models (\textit{Intrinsic3d} and \textit{Scene3d}).

\begin{table}[!t]
\renewcommand{\arraystretch}{1.3}
\caption{Running time (minutes) of different methods on different real datasets.}
\label{tab:runningtime}
	\centering
    \scalebox{0.88}{
		\begin{tabular}{lccccc}
			\hline
			{Datasets} & \tabincell{c}{G2LTex} &
			\tabincell{c}{ATO} & 
			\tabincell{c}{Intrinsic3d} & 
			\tabincell{c}{JointTG} & 
			\tabincell{c}{Ours} \\
			\hline
			Dolls (9 models) & 24.16 & 15.47 & 18.54 & 16.26 & 26.84\\
			Intrinsic3d (5 models) & 805.75 & 31.18 & 160.26 & 72.94 & 49.88\\
			Scene3d (12 models) & 368.14 & 38.64 & 69.97 & 63.31 & 53.66\\
			Chairs (35 models) & 59.89 & 10.06 & 39.62 & 15.11 & 16.76\\
			\cellcolor{Gray} Average (all models) & \cellcolor{Gray}  \textbf{176.39} &\cellcolor{Gray}  \textbf{18.21} & \cellcolor{Gray} \textbf{52.37} &\cellcolor{Gray}  \textbf{29.50} & \cellcolor{Gray} \textbf{28.22}\\
			\hline
		\end{tabular}
	}
	\normalsize
\end{table}

\subsection{Ablation Studies} 
\label{sec:abl}
\begin{figure}[!t]
\centering
\includegraphics[width=0.98\linewidth]{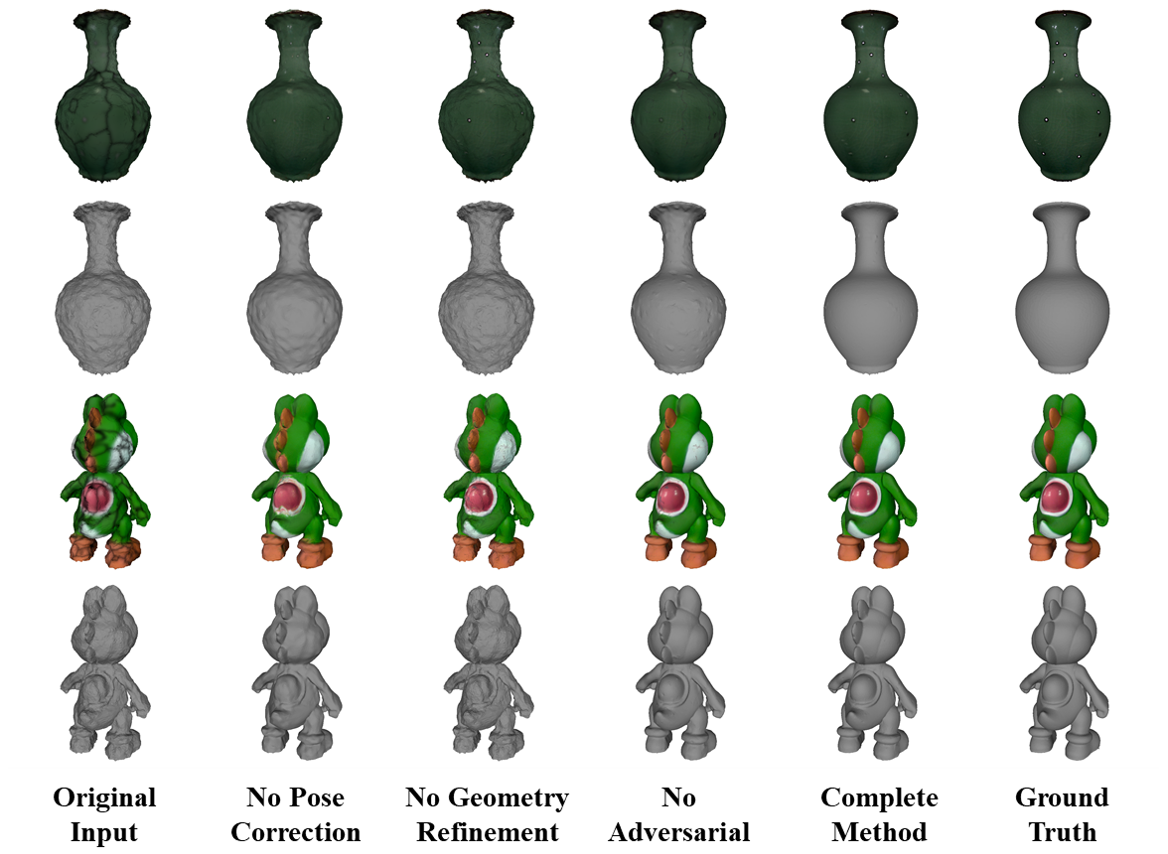}
\caption{Comparison of the texture and geometry in the ablation studies.}
\label{fig:ablation}
\end{figure}

We investigated the effectiveness of each major component in our method using the synthetic dataset. Firstly, to validate the effect of joint optimization, we conducted an experiment by removing camera pose correction, geometry refinement, or adversarial loss of texture optimization from our joint optimization framework. Without the camera pose correction, the rendered image was misaligned with the RGB-D inputs. Therefore, the reconstruction objectives were wrongly calculated and caused a severe performance drop, as shown in Table \ref{tab:ablation} and Figure~\ref{fig:ablation}. Removing geometry refinement was less significant than removing the camera-pose correction, since the influence of geometry errors is more local. Still, the performance was worse than that of the complete method. Compared to the case without adversarial loss, our complete method can produce more realistic texture details. This ablation study showed that the camera pose, geometry, and texture affect each other, and thus a joint optimization is necessary.

\begin{figure}[!t]
\centering
\includegraphics[width=0.95\linewidth]{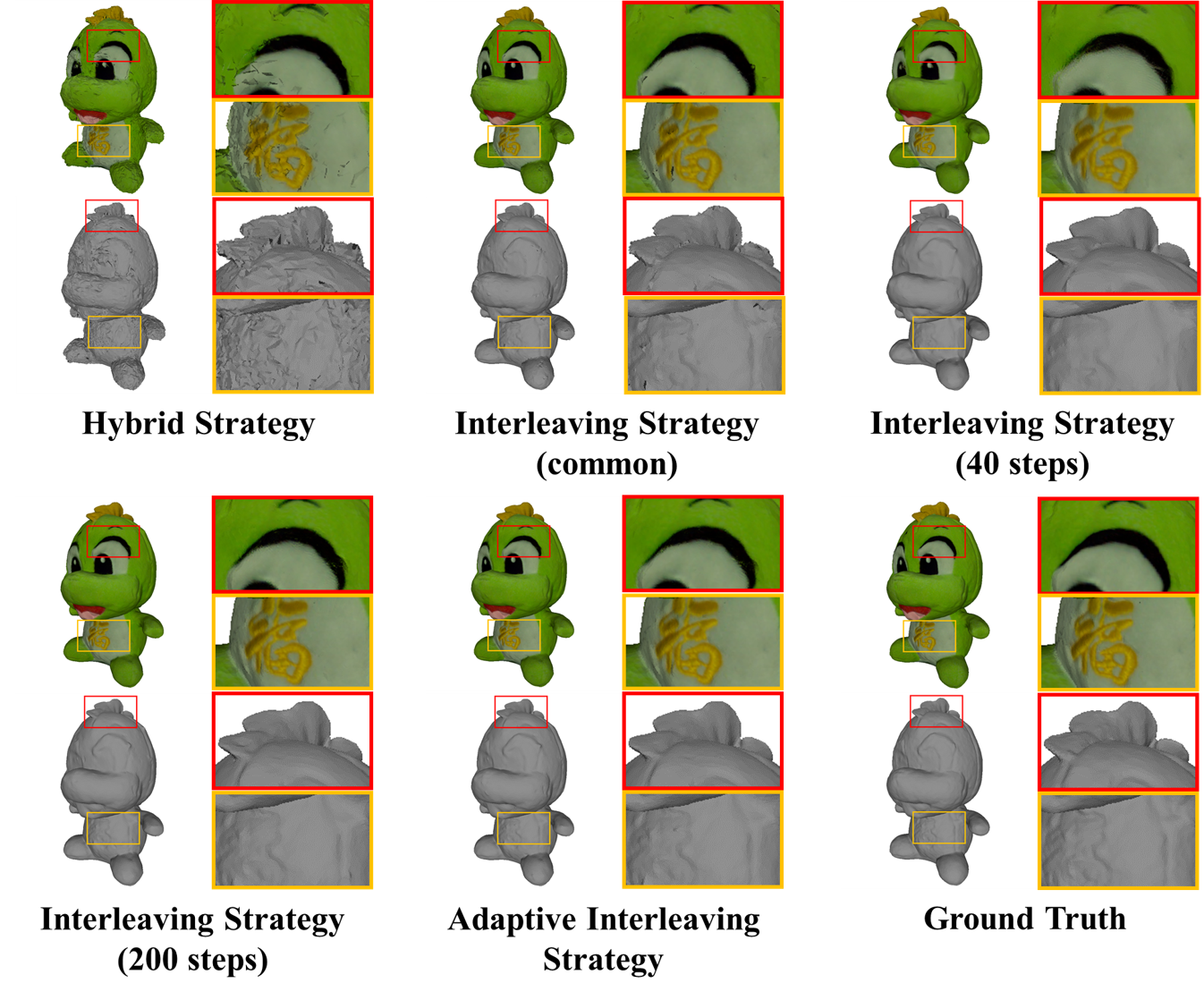}
\caption{Comparison of the texture and geometry produced by different strategies.}
\label{fig:ablation_strategy1}
\end{figure}

\begin{figure}[!t]
\centering
\includegraphics[width=1\linewidth]{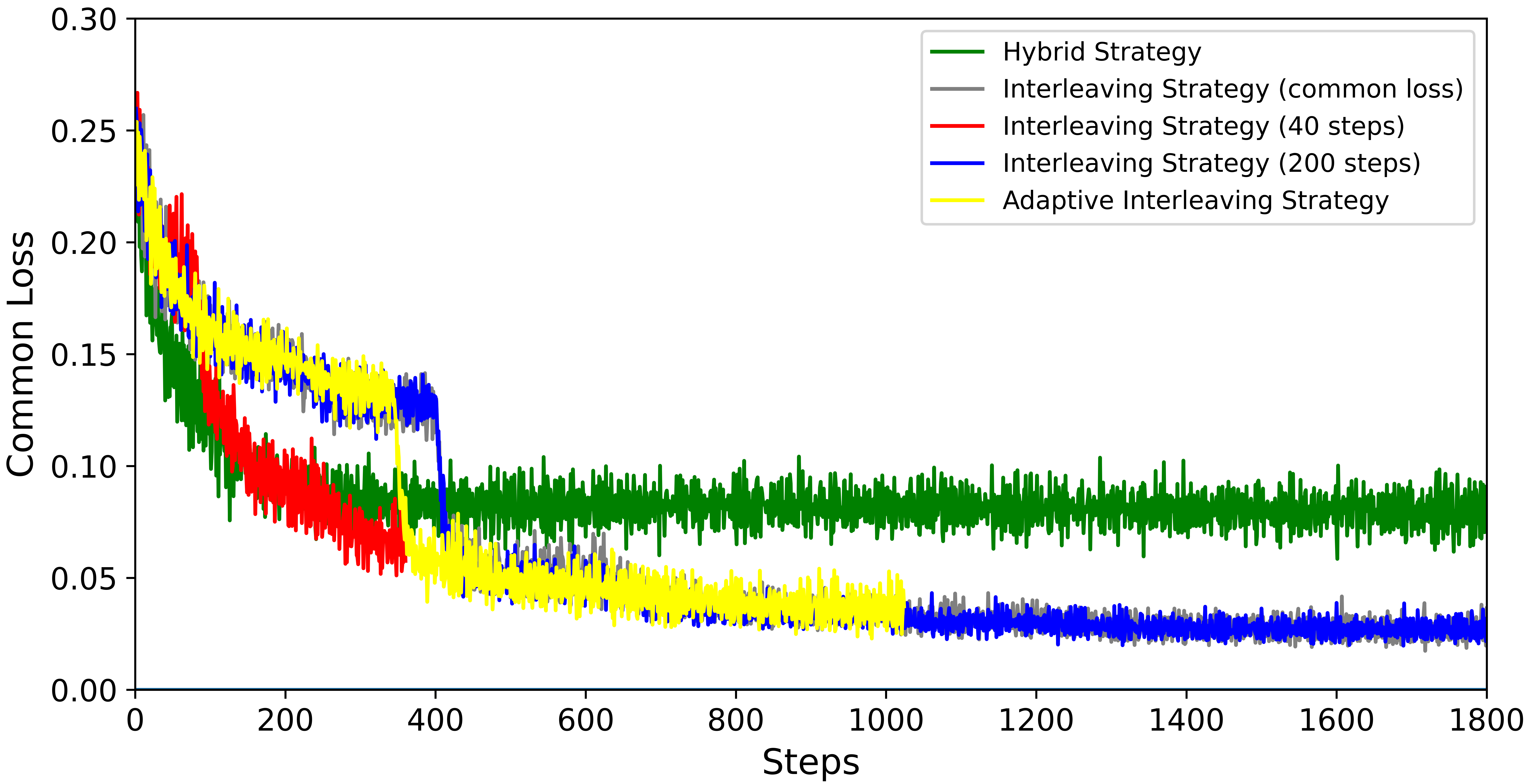}
\caption{Common loss produced by different strategies. The interleaving strategy (40 steps) finished at 360\textit{th} step, and our adaptive interleaving strategy automatically terminated at around 1050\textit{th} step.}
\label{fig:curve}
\end{figure}

To illustrate the effectiveness and superiority of our adaptive interleaving strategy, we compared it to a hybrid strategy in which texture color, mesh vertices, and camera parameters are updated together in each single iteration. The hybrid strategy was implemented based on common loss, since the Laplacian and adversarial loss cannot be added into the hybrid strategy. To be fair, we compared it to an interleaving strategy (common) without the Laplacian and adversarial loss. To further investigate the performance of our adaptive algorithm, we conducted experiments by removing the adaptive strategy and replacing it with fixed internal iteration numbers: 40 steps and 200 steps.
The qualitative comparison and convergence curves of common loss are shown in Figure \ref{fig:ablation_strategy1} and Figure \ref{fig:curve}, respectively. The common loss was averaged under a batch view in each step for all strategies, hence there are slight vibration on the curves.
The hybrid strategy led to an unstable convergence, and found it hard to reach a lower convergence position, while other interleaving strategies can overcome this problem by optimizing texture, geometry, and poses in an iterative manner. However, the internal iteration number of interleaving strategies is an important hyper-parameter, since the use of a small number of iterations may result in inadequate optimization (Figure \ref{fig:curve} (red)), while a large one can be unnecessarily time-consuming (Figure \ref{fig:curve} (blue)). Thus, it is important to dynamically select a suitable internal iteration number via the adaptive algorithm.

We report the quantitative comparison among different strategies in Table \ref{tab:abl2}. These results show that our adaptive interleaving strategy had a similar performance as the 200-step interleaving strategy, but was much more time-efficient. Compared to the hybrid strategy and the 40-step interleaving strategy, our strategy was more effective and produced better optimized results. To sum up, the proposed adaptive interleaving strategy could better balance the performance and efficiency of the optimization algorithm.

\begin{table*}[!t]
\renewcommand{\arraystretch}{1.3}
\caption{Quantitative comparison between our adaptive interleaving strategy and other strategies on synthetic datasets.}
\label{tab:abl2}
	\centering
    \scalebox{0.99}{
		\begin{tabular}{lccccc}
			\hline
			{Methods} & \tabincell{c}{Hybrid Strategy} &
			\tabincell{c}{Interleaving Strategy\\ (common)} & 
			\tabincell{c}{Interleaving Strategy\\ (40 steps)} & 
			\tabincell{c}{Interleaving Strategy\\ (200 steps)} & 
			\tabincell{c}{Adaptive Interleaving\\ Strategy} \\
			\hline
			PSNR$\uparrow$ & 28.683 & 33.187 & 29.926 & 33.570 & 33.330\\
			SSIM$\uparrow$ & 0.942 & 0.967 & 0.945 & 0.967  & 0.964\\
			Perceptual$\downarrow$ & 0.054 & 0.023 & 0.039 & 0.021 & 0.023\\
			Time (minutes)$\downarrow$ & 62.589 & 78.182 & 23.872 & 120.393 & 72.283\\
			\hline
		\end{tabular}
	}
	\normalsize
\end{table*}

\begin{table}[!t]
\renewcommand{\arraystretch}{1.3}
\caption{Quantitative comparison between our complete method and the methods without camera pose correction, geometry refinement, and adversarial loss, respectively.}
\label{tab:ablation}
	\centering
    \scalebox{0.80}{
		\begin{tabular}{lcccc}
			\hline
			{Methods} & \tabincell{c}{No pose \\ correction} & \tabincell{c}{No geometry \\ refinement} & \tabincell{c}{No adversarial \\ loss} & \tabincell{c}{Complete \\ method} \\
			\hline
			PSNR$\uparrow$ & 25.515 & 29.666 & 32.763 & \textbf{33.330}\\
			SSIM$\uparrow$ & 0.910 & 0.944 & 0.964 & \textbf{0.964}\\
			Perceptual$\downarrow$ & 0.094 & 0.047 & 0.023 & \textbf{0.023}\\
			Hausdorff$\downarrow$ ($\times10^{-2}$) & 0.422 & 0.517 & 0.165 & \textbf{0.159}\\
			\hline
		\end{tabular}
	}
	\normalsize
\end{table}

\subsection{Extension: Optimization with RGB Data}
\label{sec:exten}
Although our adaptive joint optimization method was originally implemented based on RGB-D data, it is readily applied to RGB data, by removing the depth loss from the framework. Although there is still a certain gap in the geometry details compared with our complete framework, the version without depth loss can produce reasonably clear texture and a plausible geometric model, as shown in Figure \ref{subfig_RGB}.
A quantitative comparison of these two pipelines performed on our \textit{Doll} dataset, as shown in Table \ref{tab:extension_depth}, indicated that our pipeline produced acceptable performance with only RGB data.
Our method can be extended to joint optimization with RGB or RGB-D data, while the depth information can contribute to a better geometric details.

\begin{table}[!t]
\renewcommand{\arraystretch}{1.3}
\caption{Quantitative comparison of optimization results produced by our method w/ and w/o depth information on our \textit{Doll} dataset.}
\label{tab:extension_depth}
	\centering
    \scalebox{0.99}{
		\begin{tabular}{l ccc}
			\hline
			{Methods} & PSNR$\uparrow$ & SSIM$\uparrow$ & Perceptual$\downarrow$ \\
			\hline
			Ours (RGB) & 24.360 & 0.944 & 0.045\\
			Ours (RGB-D) & 25.185 & 0.949 & 0.042\\
			\hline
		\end{tabular}
	}
\end{table}

\subsection{User Study}
\label{sec:user}
\begin{figure}[!t]
\centering
\includegraphics[width=1\linewidth]{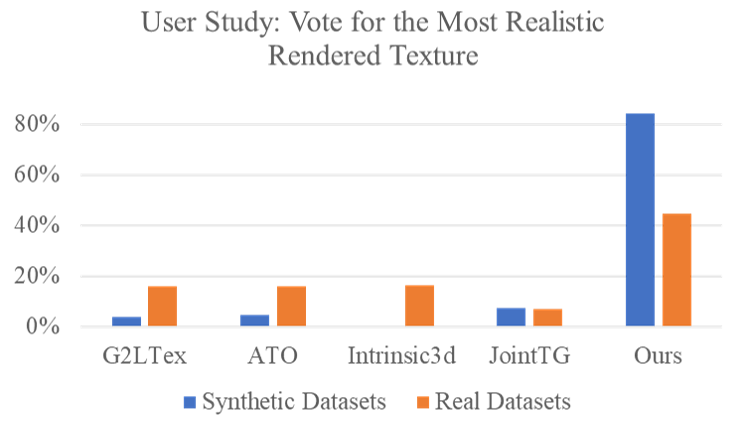}
\caption{User study. We asked participants to vote for the most photo realistic 3D model optimized using different methods.}
\label{fig:user}
\end{figure}

In order to further evaluate the quality of the optimized results on both synthetic and real data, we performed a study in which we asked users to vote for the most visually realistic rendered image, as shown in Figure \ref{fig:user}. A total of 138 participants were asked to vote for the best rendered result on both synthetic and real datasets. During the survey process, we showed the front rendering of each model, and there was no time limit for participants to vote. For some real datasets with low noise levels, it was sometimes difficult for individuals to distinguish between the rendered images produced by the different methods. Nevertheless, images produced by the method proposed in this paper was still preferred over those produced by the other methods.

\section{Conclusions}
We proposed a novel adaptively joint optimization method for 3D reconstruction based on differentiable rendering. This method integrates the optimization of camera pose, geometry, and texture into a unified framework, and achieved better performance in recovering both fine-scale geometry and high-fidelity texture, compared with state-of-the-art methods. We also conducted ablation studies to demonstrate the effectiveness and  superiority of each major component and the adaptive interleaving strategy.
In the future, approaches to migrating our framework from iterative optimization to a feed-forward network, to improve performance would be a direction worth exploring.


\ifCLASSOPTIONcaptionsoff
  \newpage
\fi


\bibliographystyle{IEEEtran}
\bibliography{IEEEabrv,ref}

%




\end{document}